\documentclass[10pt,onecolumn,letterpaper]{article}
\usepackage{cvpr}  
\makeatletter
\@namedef{ver@everyshi.sty}{}
\makeatother
\usepackage{tikz}

\usepackage{graphicx}

\usepackage{tikz}
\usepackage{comment}
\usepackage{amsmath,amssymb} 
\usepackage{color}

\usepackage[accsupp]{axessibility}  

\usepackage{multirow}
\usepackage{lipsum}  
\usepackage{bbm}
\newcommand{\sign}[1]{\mathrm{sgn}(#1)}

\begin{document}

\title{Discrete-Constrained Regression 
for Local Counting Models}

\author{Haipeng Xiong\\ 
National University of Singapore, Singapore\\
{\tt\small haipeng@comp.nus.edu.sg}
\and
Angela Yao\\
National University of Singapore, Singapore\\
{\tt\small ayao@comp.nus.edu.sg}
}
\maketitle

\begin{abstract}
Local counts, or the number of objects in a local area, is a continuous value by nature. Yet recent state-of-the-art methods show that formulating counting as a classification task performs better than regression. Through a series of experiments on carefully controlled synthetic data, we show that this counter-intuitive result is caused by imprecise ground truth local counts. Factors such as biased dot annotations and incorrectly matched Gaussian kernels used to generate ground truth counts introduce deviations from the true local counts. Standard continuous regression is highly sensitive to these errors, explaining the performance gap between classification and regression. To mitigate the sensitivity, we loosen the regression formulation from a continuous scale to a discrete ordering and propose a novel discrete-constrained (DC) regression.  Applied to crowd counting, DC-regression is more accurate than both classification and standard regression on three public benchmarks.  A similar advantage also holds for the age estimation task, verifying the overall effectiveness of DC-regression. 

\end{abstract}

\section{Introduction}

Image-based counting of objects such as people~\cite{zhang2015cross,CSRNet_2018_CVPR}, vehicles~\cite{onoro2016towards} and cells~\cite{lempitsky2010learning} can be modelled either as a classification or a regression problem.  Since \emph{local counts} or \emph{local densities} are continuous, ordered values, they should naturally be regressed. Yet surprisingly, recent works have shown that formulating local count prediction as a classification problem is more accurate~\cite{xiong2019open,liu2019counting}. 

The preference for using classification to solve regression problems arises in several areas of computer vision, ranging from depth estimation~\cite{li2018deep} to human pose estimation~\cite{newell2016stacked,xiao2018simple}. The underlying reason is usually task-specific. For depth estimation, classification helps to handle the extreme dynamic range of depth that may occur; this is especially prominent in mixed indoor and outdoor scenes~\cite{li2018deep}. For pose estimation, classification allows for dense spatial supervision, which is believed to be more beneficial for learning~\cite{gu2022dive}.  

For counting tasks, classification outperforms regression when counting in a closed set range~\cite{xiong2019open}. To find out why classification performs better than regression in certain counting tasks, we conduct a series of experiments on synthetic data. In this paper, we focus on local counting models~\cite{lempitsky2010learning,xiong2019open,liu2019counting,liu2020adaptive,xiong2019tasselnetv2} that predict local counts,~\ie~the number of objects within the local image patches. Through careful investigation, we trace the advantages of classification back to the imprecise generation of ground truth local count.   

\begin{figure}[!t]
    \centering
    \includegraphics[width=0.95\linewidth]{ 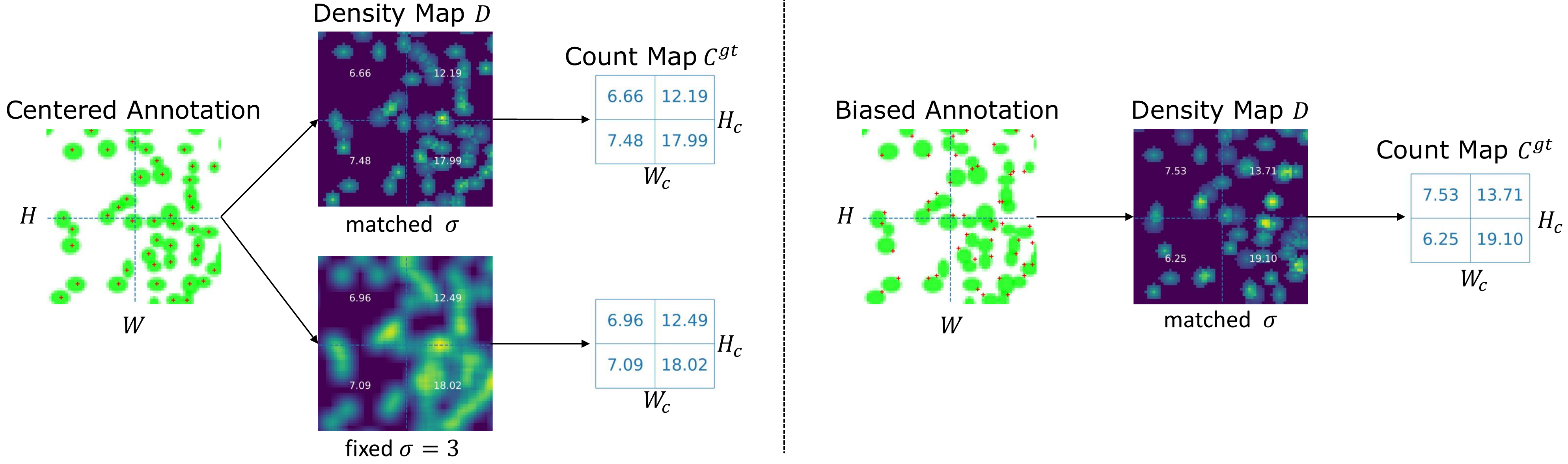}
    \caption{Visualization of errors in the ground truth count maps. Left: A sample $64 \times 64$ input image with $4$ local patches with centered annotations. (Top) Gaussian kernels dynamically match the object size vs. (Bottom) Gaussian kernels with a fixed standard deviation of $\sigma=3$. Right: The same input image with biased dot annotations, generated from Gaussians matching the object size. Note that the sum of the local counts differs slightly as some kernels go beyond the local image borders.}
    \label{fig:local_count_generate}
\end{figure}

In object counting, annotators mark each object of interest with a single dot (see Fig.~\ref{fig:local_count_generate}).  The \emph{local count}, or the number of objects within a specified area, can simply be defined as the number of dot annotations within that area if all the objects are wholly contained. However, fractional local counts arise when there are partial objects. The estimate of fraction local counts can be imprecise, especially if the dot annotations are not aligned with the object in the image. The imprecision is worsened by the use of an intermediate \emph{density map} estimated by convolving the dot annotations with Gaussian kernels.  An accurate density estimate requires the Gaussian kernels' standard deviation to match the true object size.  Yet, size information is often not annotated or known in advance. Given that current methods~\cite{xiong2019open,xiong2019tasselnetv2,liu2020adaptive} generate ground truth counts by integrating over local areas in the density map, it becomes clear that many factors of imprecision are at play and that the ground truth local counts will deviate from the true local counts.

In this work, we investigate the effects of partial objects, dot annotation position and Gaussian kernel size on the classification versus regression task setting. To that end, we create a synthetic dataset containing images of cells to carefully control and study these properties.  Our findings show that regression performance is comparable to classification under ideal settings,~\ie~objects are wholly contained within a local area, dot annotations are centered on the objects and Gaussian kernels are correctly matched. The lower accuracy arises only when the ground truth labels deviate from the true local count.  

Based on these findings, we speculate that the poor performance of regression can be attributed to the naive formulation as a \emph{continuous} regression problem.  Enforcing a standard L1 loss penalizes models that do not (over-) fit to the errors present in the ground truth counts.  By constraining the regression to only \emph{discrete} counts, however, we can buffer against some of the imprecision in the ground truth as classification does. As such, we propose a discrete-constrained (DC) regression model for crowd counting. Similar to classification, DC-regression benefits from discretizing the target space while retaining an ordered output space of numerical counts rather than an order-agnostic class index. The numeric output enables a comparison between summed local counts and a global count. We also propose a global count loss regularizer, which mitigates some of the discretization errors when converting the target space from continuous to discrete. 

DC-regression is more accurate than both classification and standard regression in crowd counting. It is also applicable to other discrete regression tasks such as age estimation, and we verified its effectiveness in our experiments. In summary, the contributions of this paper are:

\begin{enumerate}
    \item A novel discrete-constrained (DC) regression model that benefits from the discretization of classification while retaining the ordering and numerical output space of regression. 
    \item A series of experiments on controlled synthetic data indicated that the imprecise local counts emerged during the annotation and ground truth generation process account for the advantage of classification over regression.
    \item DC-regression, coupled with a global count loss, outperforms both classification and regression on three crowd-counting datasets. It has significant improvements over state-of-the-art approaches when inserted into an advanced local count model like S-DCNet~\cite{xiong2019open}.
    \item Verified that DC-regression is also applicable to the age estimation task.
\end{enumerate}

\section{Related Work}

\textbf{Counting By Regression.} The goal of object counting is to predict the number of visually present objects in images. Given that counts have an ordering, it is naturally modelled as a regression problem. Two commonly adopted regression targets are density maps~\cite{MCNN_2016_CVPR,CSRNet_2018_CVPR,sindagi2019multi,shi2019counting} and local object counts~\cite{paul2017count,xiong2019tasselnetv2,liu2020adaptive}. Density maps were first proposed by~\cite{lempitsky2010learning}; they are a dense target proportional to the spatial distribution of the objects. Zhang~\etal~\cite{zhang2015cross} first adopted a deep network to regress density maps and most regression methods~\cite{MCNN_2016_CVPR,CSRNet_2018_CVPR,sindagi2019multi,shi2019counting} followed a similar approach. Local count is another learning target of deep regression networks.  Local count methods~\cite{paul2017count,xiong2019tasselnetv2,liu2020adaptive} first divide images into local patches, then predict object numbers of each local patch separately. 

However, both learning targets could be imprecise if only dot annotations are provided for object counting. Specifically, ground truth density maps could be affected by biased dot annotations and mismatched Gaussian kernels (as shown in Fig~\ref{fig:local_count_generate}). Local counts can be even more imprecise as they are obtained by integrating over local areas in the density map. Optimum transport~\cite{peyre2019computational} has been adopted~\cite{wang2020DMCount,wan2021generalized,ma2021learning} to account for possible errors in ground truth density maps with extra optimization procedures. Different from them, we tackle errors in ground truth count maps using discrete constraints.\\

\noindent \textbf{Counting By Classification.} Instead of regressing local counts, counting by classification methods model local counts as different classes~\cite{xiong2019open,liu2019counting,wang2021uniformity}. Liu~\etal~\cite{liu2019counting} first divided the count ranges into discrete intervals and then predicted the interval index with a classifier. The final count value was chosen as the median value of the predicted interval. Xiong~\etal~\cite{xiong2019open} adopted a classifier to model a closed-set range of counts, and generalized it to the open-set range via spatial divide-and-conquer. Based on these works~\cite{xiong2019open,liu2019counting,wang2021uniformity}, it appears that classification works better than regression for object counting.\\

\noindent \textbf{Classification vs. Regression.} Other than object counting, classification has also shown to perform better than regression for specific tasks in depth estimation~\cite{li2018deep}, human pose estimation~\cite{newell2016stacked,xiao2018simple} and age estimation~\cite{zhang2017quantifying}. In this work, we show experimentally that classification is better than regression when ground truth count maps are imperfect. We also find that regression could be improved by adopting discrete constraints of local counts similar to classification.

\section{Method}
\subsection{Preliminaries} \label{subsec:preliminaries}

Suppose we are given an image $I \in \mathbb{R}^{H \times W}$ with $T$ dot annotations $(x_t,y_t)$, $t=\{1,\dots,T\}$.  Using the dot annotation map $D^0= \sum_{t=1}^{T} \delta(x_t,y_t)$, where $\delta(x_t,y_t)$ denotes a Dirac function centered at $(x_t,y_t)$, we can generate a density map $D$ by convolving $D^0$ with a Gaussian kernel $G_{\sigma}$, with a standard deviation of $\sigma$\footnotemark\footnotetext{This work assumes a kernel size of $4\sigma$ and use the terms `size' and `$\sigma$' interchangeably.}:
\begin{equation} \label{eq:density}
    D = \sum_{t=1}^{T} \delta(x_t,y_t)*G_{\sigma}
\end{equation}
Summing the local density in the $P_h \times P_w$ non-overlapping window, the ground truth count maps $C\!\in\! \mathbb{R}^{H_c \times W_c}$, where $H_c\!=\!H/P_H$ and $W_c\!=\!W/P_w$ is defined as:
\begin{equation} \label{eq:count}
    C(j,k) =   \sum_{h=j\times P_h}^{(j+1)\times P_h} \sum_{w=k\times P_w}^{(k+1)\times P_w}  D(h,w).
\end{equation}

A local count model predicts a corresponding local count map $\hat{C}\in \mathbb{R}^{H_c \times W_c}$. In practice, $C$ is imprecise when dot annotations $(x_t,y_t)$ are biased or Gaussian kernels $G_\sigma$ are mismatched. Consider an object near the border of the local area, if the dot annotations are off-center or Gaussian kernels have too large a standard deviation, then some portions of the density may shift to other nearby local areas, leading to imprecision in Eqs.~\eqref{eq:density} and~\eqref{eq:count}.

The ground truth and estimated count for image $I$ can be estimated by summing the respective local count maps: 
\begin{equation}
c =\sum_
{j,k}^{H_c,W_c} 
C(j,k), \qquad 
 \hat{c} =\sum_
 {j,k}
 ^{H_c,W_c} 
 \hat{C}(j,k),
\end{equation}
where $j$ and $k$ are row and column-wise indices of the count map.  
The error of the local count is defined as the difference between the ground truth and predicted local counts. The error of local count $E \in \mathbb{R}^{H_c \times W_c}$ and global count $e$ is computed as:
 \begin{equation}
    e = c-\hat{c} = \sum_{j,k}
    ^{H_c, W_c} 
    E (j,k), \quad \text{where} \quad     E = C-\hat{C}.
\end{equation}

A typical regression-based local count model would use a CNN backbone and add a dedicated regression head. The final model would take image $I$ as input, output the local count map $\hat{C}$ and be trained with an L1 loss:
\begin{equation} \label{eq:Lreg}
    L_{reg} = \frac{1}{H_cW_c}  \sum_{j,k}^{H_c,W_c}|E(j,k)|.
\end{equation}
Note that eq.~\eqref{eq:Lreg} denotes the loss for one image; this loss is averaged over all images in the batch during training.

\subsection{Discrete-Constrained Regression}
But as we argue that ground truth count maps are imprecise and may contain error $\epsilon$,~\ie~ $C = C^{true} + \epsilon$, where $C^{true}$ is the count map generated with ideal density map.  The regression loss in Eq.~\eqref{eq:Lreg} then becomes
\begin{equation} \label{eq:Lreg_error}
    L_{reg} 
    =\frac{1}{H_cW_c} \sum_{j,k}^{H_c,W_c}  |E^{true}(j,k)+ \epsilon(j,k)|, \quad \text{where} \quad E^{true} = C^{true}-\hat{C}.
\end{equation}
\noindent When $|E(j,k)|>=|\epsilon(j,k)|$, the sign of the observed error and true error remains the same,~\ie~ $\sign{E_i(j,k)} = \sign{E^{true}(j,k)}$, where ``$\sign{\cdot}$'' denotes the sign function. As such, the gradient will also be the same,~\ie~
\begin{equation}
    \frac{\partial |E(j,k)|}{\partial \hat{C}(j,k)} = \frac{\partial |E^{true}(j,k)|}{\partial \hat{C}(j,k)} =-\sign{E(j,k)}.
\end{equation}

To account for the $\epsilon$, we opt to partition the target space into discrete intervals.  Specifically, we follow a classification setup and a range $[V_{min},V_{max}]$ into $N+1$ intervals
$\{V_0\}$, $(V_0,V_1]$, $(V_1,V_2]$, $...$, $(V_{N-1},V_{N}]$, where $V_0=V_{min}$ and $V_{N}=V_{max}$. In counting datasets, $V_{min}=0$ and $V_{max}$ is the maximum count value in the training set. A count value $c$ that falls into  $[V_{i-1},V_{i}]$, where $(i=1,2,...,N)$, would be associated with the $i$-th interval. As such, the loss is considered to be correct when the prediction $\hat{C}(j,k)$ is outside the range $[V_{G(j,k)},V_{G(j,k)+1})$, where $G(j,k)$ is the index of $C(j,k)$. 

We formulate the discrete-constrained loss $L_{dc}$ as:
\begin{equation} \label{eq:ldc}
    L_{dc} = \frac{\sum_{j,k}^{H_c,W_c} S(j,k)\times |E(j,k)|}{ \sum_{j,k}^{H_c,W_c} S(j,k)} 
\end{equation}
where the mask $S =  1-\mathbbm{1}\{V_{G}<\hat{C}<=V_{G+1}\}$, $\mathbbm{1}\{\}$ is the indicator function and $G$ is the index of $C$. $S$ only selects the samples that are predicted outside the intervals to compute loss, which ensures the gradient directions to be correct.

\subsection{Global Count Loss $L_{\text{gc}}$} \label{subsec:L_gc_define}
The overall aim of counting is to estimate a global count from summing all the local counts.  Even if all the local counts are correctly predicted,~\ie~$L_{dc}=0$, there may still be a gap between their sum and the GT global count, precisely due to the quantization. We could use a global count loss $L_{\text{gc}}$ to decrease the quantization errors. Naively, an $L_1$ loss could be applied to the global count, \ie
\begin{equation} \label{eq:L_count}
    L_{\text{c}} = \frac{|e|}{ H_c W_c},
\end{equation}
However, $L_{\text{c}}$ is problematic as it produces the same gradient for all the local counts, regardless of over- or under-estimation. We therefore improve $L_{\text{c}}$ to $L_{\text{bias}}^{0}$ by selecting local patches that have the same trend as global error,~\ie, considering only over-estimated patches if the global count is an over-estimate and and same for under-estimation. This prevents the wrong gradient for patches with the opposite sign of global error $e$. $L_{\text{bias}}^{0}$ can be defined as

\begin{equation}
    L_{\text{bias}}^{0}= \frac{\sum_{j,k}^{H_c,W_c}  S^{a}(j,k)\times |E|}{ \sum_{j,k}^{H_c,W_c} S^{a}(j,k)},
\end{equation}
where $S^{a}=\mathbbm{1}\{\sign{ e}\times E(j,k)>0\}$, and $\sign$ denotes the sign function. Among patches with $S^{a}=1$, we further discard those patches with smaller errors that compensate the error of patches with $S^{a}=0$. We achieve this by introducing a threshold $\lambda$, which satisfies
\begin{equation}
\sum_{j,k}^{H_c,W_c}   S^{a}(j,k) \times \mathbbm{1}\{\sign{e}\times E(j,k)>=\lambda\} \times|E(j,k)| = \sum_{j,k}^{H_c,W_c}|E(j,k)|
\end{equation}
and select patches via $S^{m}(j,k) = S^{a}(j,k) \times \mathbbm{1}\{\sign{e}\times E(j,k)>=\lambda\}$. Now, an adjusted loss $L_{\text{bias}}^{\lambda}$ can be computed as
\begin{equation} \label{eq:L_bias_lambda}
    L_{\text{bias}}^{\lambda} = \frac{\sum_{j,k}^{H_c,W_c}  S^{m}(j,k)\times |E(j,k)|}{ \sum_{j,k}^{H_c,W_c} S^{m}(j,k)}.
\end{equation}

A related work~\cite{ma2019bayesian} introduced the Bayesian loss $L_{BL}$ to constrain the integration of density map to be equal to the annotated point numbers. It also serves as a type of global count loss. We compare these variants of global count losses in the ablation studies. 

\section{A Synthetic Dataset Investigation on Local Counting}

\subsection{Data Preparation}
Inspired by~\cite{lempitsky2010learning,xiong2019open}, we create a synthetic dataset of cells to study counting-related factors in a controlled setting. Each synthetic image was fixed to $128\times 128$, which could be further subdivided into a $4\times4$ array of $32\times32$ local image patches.  Each local image patch had $0$ to $20$ oval cells, where the major and minor axis of the oval were randomly sampled from $[2,4]$ pixels. Each of the training and test sets contained $1000$ such synthetic images.

\textbf{Presence of Partial Objects} We synthesized two complete datasets. Both had the same local count distribution, but the presence of partial objects in the local $32\times 32$ patches was controlled. A comparison of the two data variants is shown in Fig.~\ref{fig:partial} (a).  When there are no partial objects, the ground truth local counts corresponds exactly to the number of dots.  With partial objects, the ground truth local count can be estimated either (i) with integer counts, based on the number of dots within the local patch, or (ii) with fractional counts by integrating (summing) the Gaussian-convolved density map using a Gaussian kernel with the same size as the cell.

\textbf{Biased Dot Annotations}
Typical real-world counting datasets~\cite{MCNN_2016_CVPR,Compose_Loss_2018_ECCV} have dot annotations.  The dots are not necessarily centered on the object and may sit anywhere within the border of the object as shown in Fig.~\ref{fig:local_count_generate}. 
The imprecise locations of the dots propagate as errors on the density maps when convolved with Gaussian kernels, which in turn creates biases in the ground truth local counts. To simulate this effect, we randomly moved the dot annotations $\Delta$ pixels in h and w directions, where $\Delta$ was uniformly sampled from $\{-a,+a\}$, where ${a=\{0,1,2,4\}}$. 

\textbf{Mismatched Gaussian Kernels} In most counting datasets~\cite{MCNN_2016_CVPR,Compose_Loss_2018_ECCV}, the sizes of the objects are not known. This makes the Gaussian-convolved density map an imprecise estimate since a fixed-sized Gaussian kernel is applied when it should be a function of the object size.  We investigate the effect of Gaussian kernel size on local counting models. We adopt $0$, $3$, $6$ as the deviations of Gaussian kernels to generate density maps and then integrated density maps to obtain local count maps. A deviation of $0$ means using the dot annotations directly. We also add a baseline where the Gaussian kernel sizes were selected according to the actual cell sizes, denoted by ``GT Size''.

\textbf{Model / Implementation Details} \label{sec:simulated}
For our local count models, we adopt all the convolutional layers in VGG16~\cite{Simonyan2014Very_VGG16} to extract feature maps, then used a regression head consisting of two $3\times3$ convolutional layers ($512$ and $1$ output channels) to map local features to local counts. The size $P_h$,$P_w$ of the local patch is $32\times32$.  An Adam optimizer was adopted for training, with a learning rate of $10^{-3}$ and a batch size of $6$. For the discrete models, we choose $20/40$ linear intervals, with an interval length of $1/0.5$, and $40$ logarithm intervals.  To evaluate, we consider the Mean Absolute Error ($\text{MAE}$), where lower $\text{MAE}$ indicates better counting performance.

\subsection{Partial Objects}

\begin{figure}[!t]
\centering
\includegraphics[width=0.47\linewidth]{ 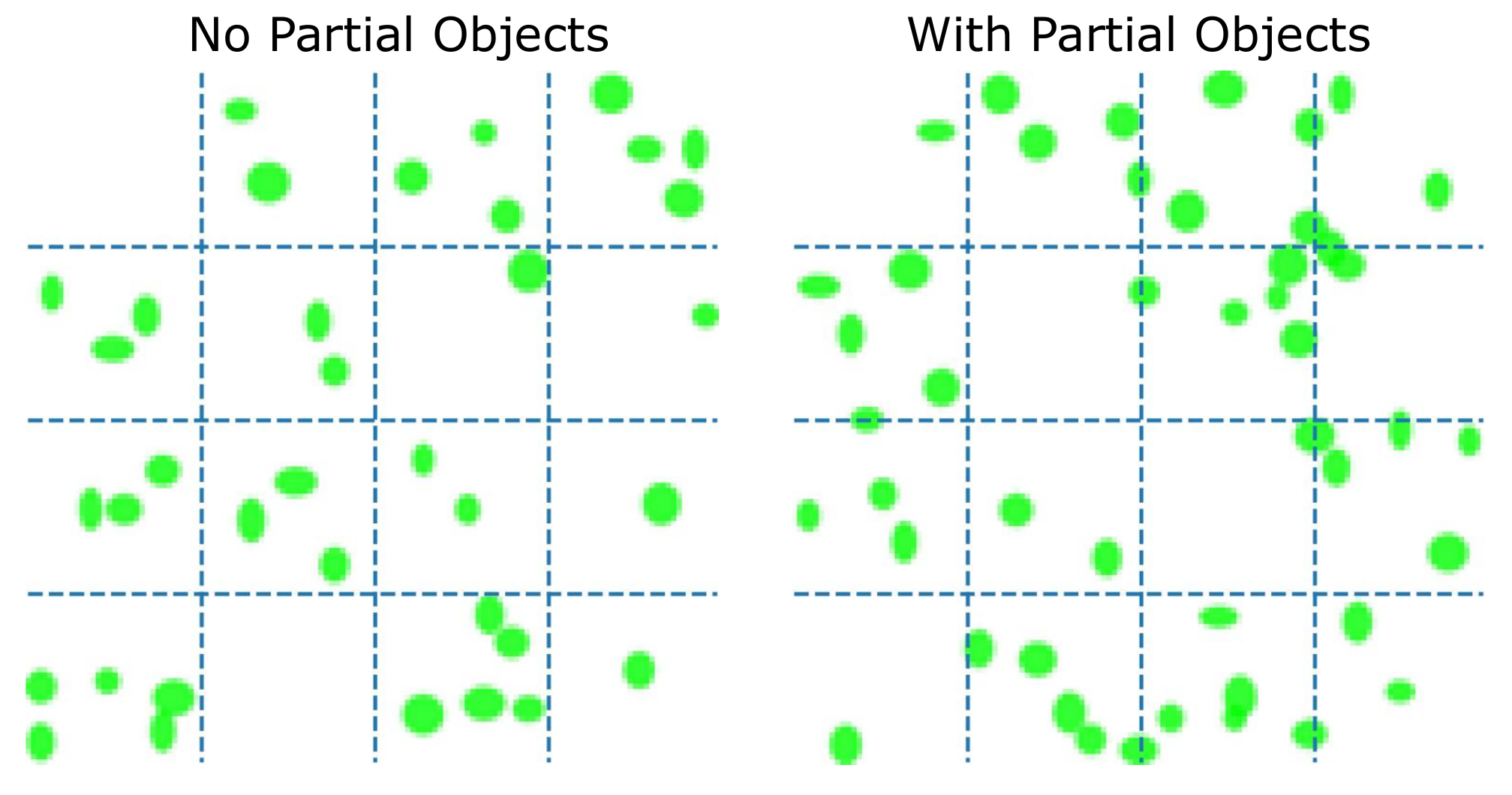}
\includegraphics[width=0.35\linewidth]{ 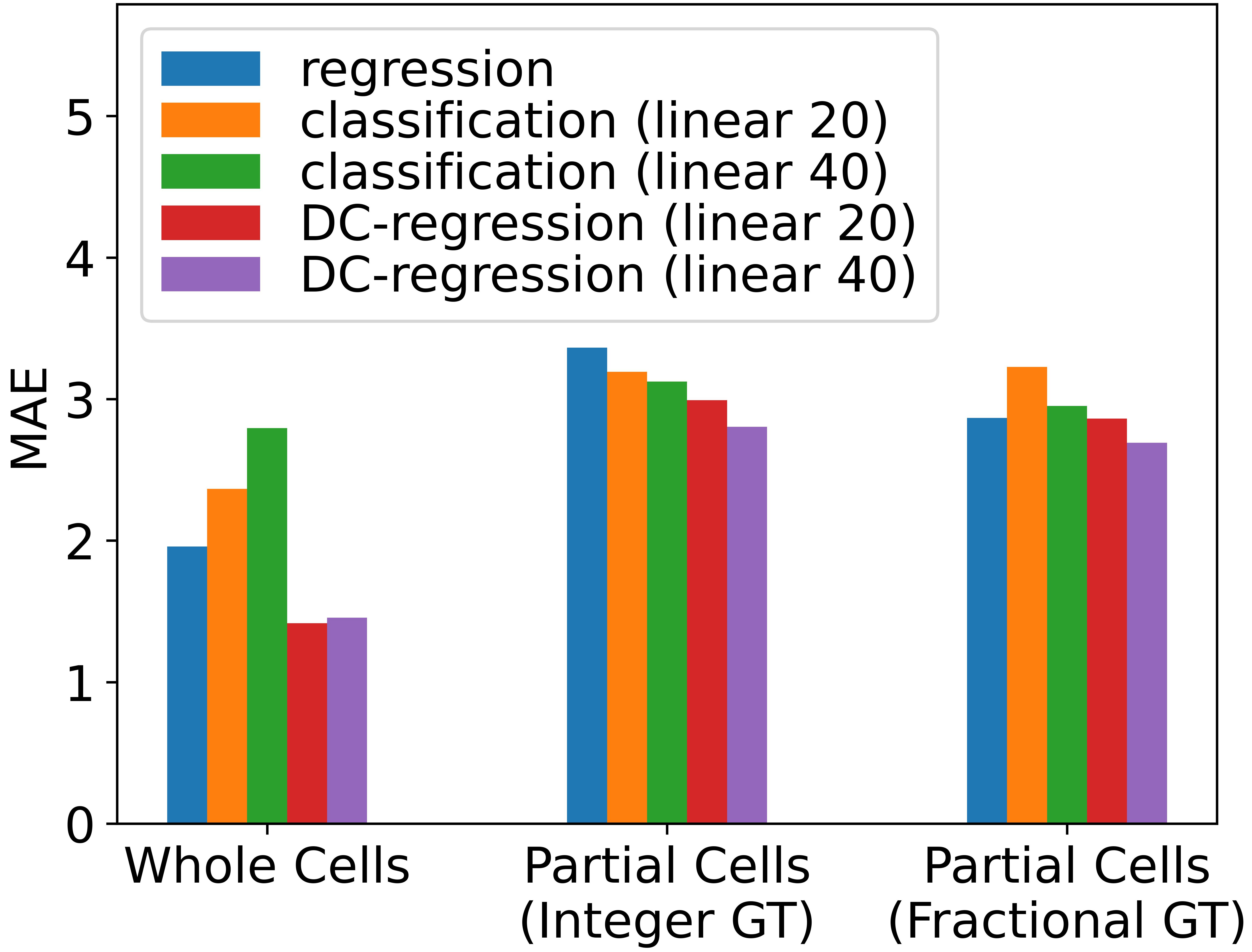}
\caption{Counting performance with respect to partial objects on the simulated cell dataset. Left: Some samples of simulated cell images without and with partial objects, respectively. Right: The counting performance of regression and discrete models. ``linear 20/40'' denotes 20 and 40 linear intervals respectively. }
\label{fig:partial}
\end{figure}

\textbf{When no partial objects} are present, the plot in Fig.~\ref{fig:partial} shows that regression outperforms classification.  When the class number increases from $20$ to $40$, counting error slightly increases for classification. This is because the local count is an integer value and increases with a step size of $1$, so half of the intervals contain no samples when there are $40$ instead of $20$ intervals. DC-regression shows the best performance. Even when the interval is $0.5$ and 40 intervals are used, the performance only drops slightly. \\

\noindent \textbf{When partial objects} are present, considering fractional counts for partial objects yields better performance. Classification shows comparable performance with regression. If fractional counts are considered for partial objects, increasing the number of intervals from $20$ to $40$ decreases the counting error.  DC-regression shows better performance than standard regression and classification.

\subsection{Incorrect Local Counts}
\begin{figure}[!t]
\centering
\includegraphics[width=0.295\linewidth]{ 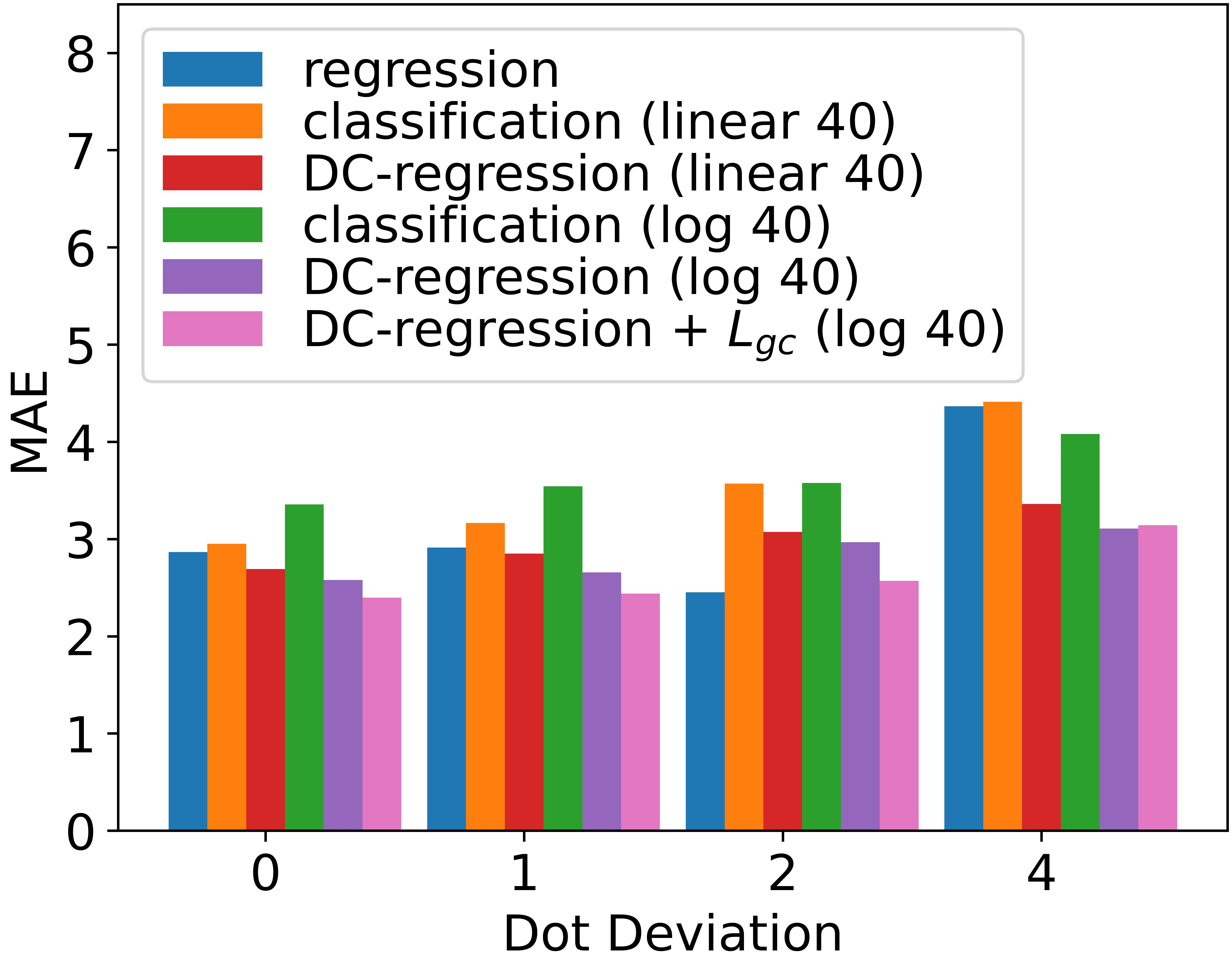}
\includegraphics[width=0.30\linewidth]{ 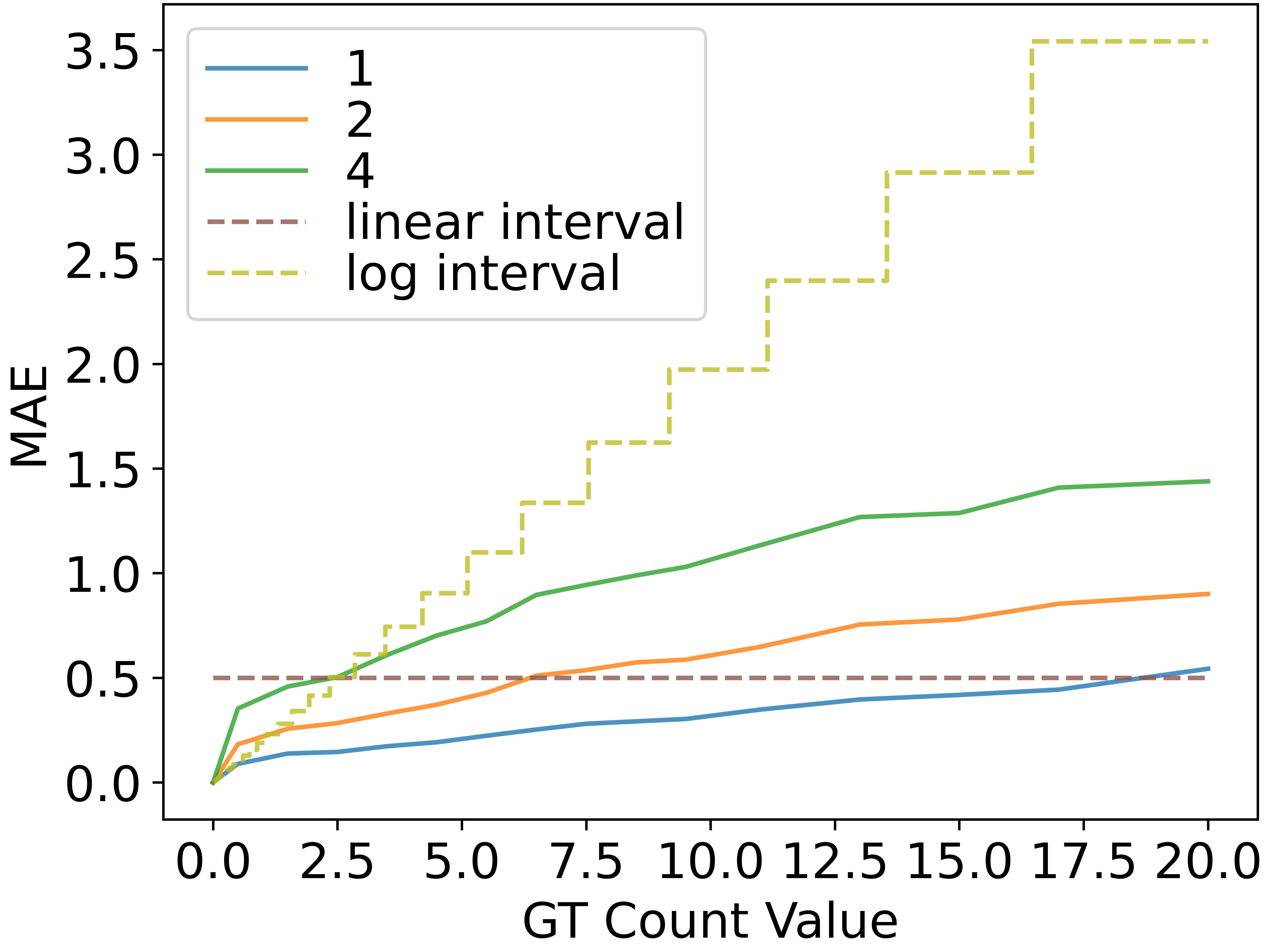}
\includegraphics[width=0.30\linewidth]{ 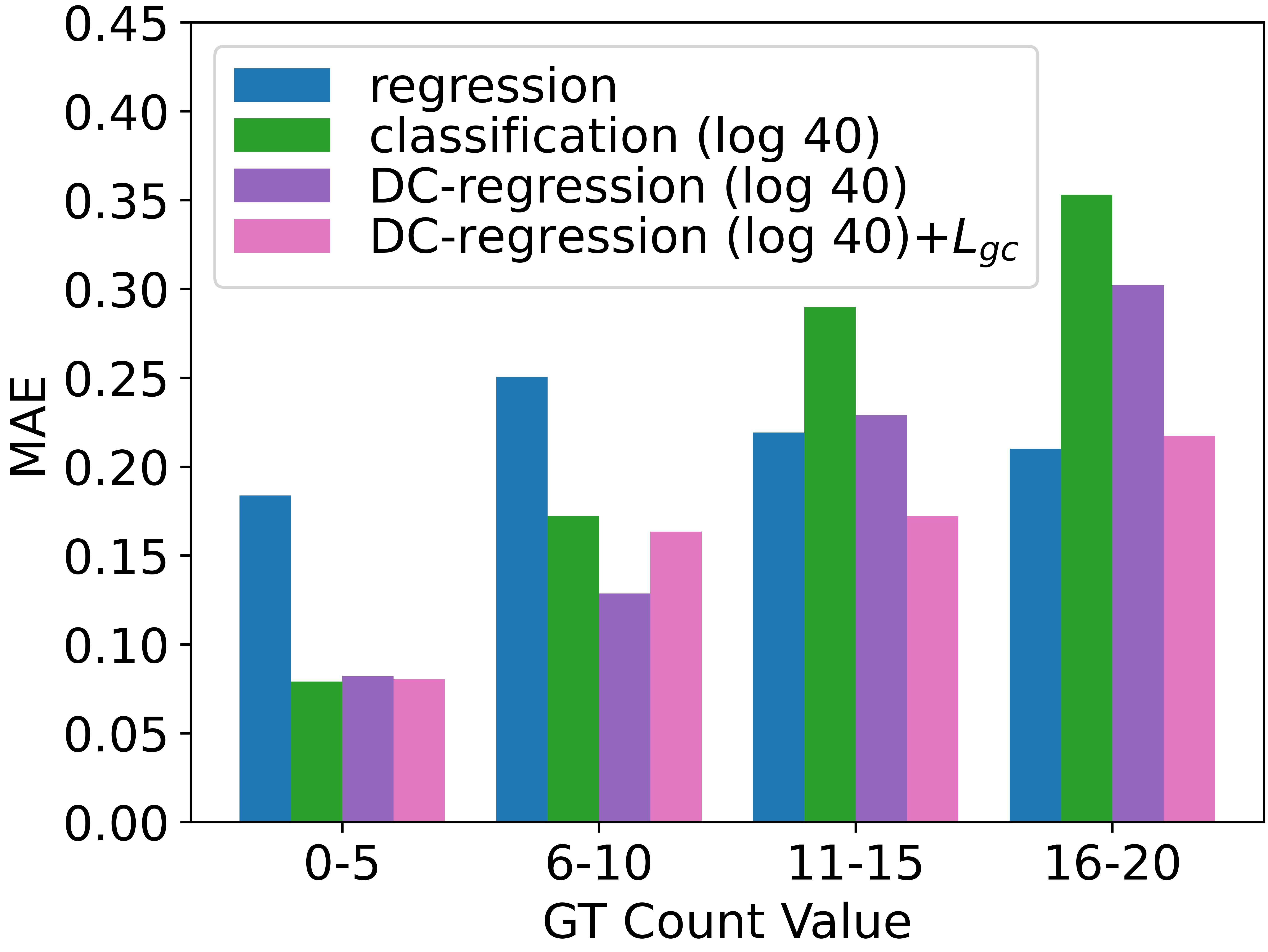}
\caption{Left: Counting results under various degrees of point annotation bias. Here $L_{\text{bias}}^{\lambda}$ is adopted as $L_{\text{gc}}$. Middle: The bias of local count value caused by point annotation bias.  Right: 
Counting error distribution w.r.t. local count 
value when dot deviation is $2$. Linear/log denotes linear and log-spaced intervals respectively, while $20/40$ denotes the number of intervals.}
\label{fig:simu_position_bias}
\end{figure}

\textbf{The impact of biased point annotations} is shown in Fig.~\ref{fig:simu_position_bias} (a). When the ground truth of local counts is imprecise, performance is highly dependent on the extent of the incorrect class label or interval. When the annotation bias is small, \eg~$1$, the ground truth error is bounded by the interval of the discrete model, but when the bias becomes large, \eg~$2$ or $4$, the ground truth error may exceed the interval length. This results in the wrong classification label being assigned.  One way to handle this is to use log-based instead of linear intervals.   In fact, the MAE increases monotonically with respect to the ground truth local count (see Fig.~\ref{fig:simu_position_bias}(b)).  This result directly implies that dense areas have higher ground truth error than sparse areas and using log-spaced intervals will outperform linearly spaced intervals.  Plots of the error with respect to the ground truth local count (see Fig.~\ref{fig:simu_position_bias} (c)) show that classification performs better than regression in sparse areas ($0\sim 5$ cells per patch) but worse in denser patches ($6-20$ cells per patch). Similarly, DC-regression shows much higher error than regression in highly dense areas ($16-20$ cells per patch), where log intervals have higher discretization error. 
However, the error of dense patches ($11\sim20$ cells per patch) decreases significantly if we add the global constraint regularizer $L_{\text{gc}}$ to the DC-regression.

\subsubsection{Mismatched Gaussian kernels} 
\begin{figure}[!t]
\centering
\includegraphics[width=0.295\linewidth]{ 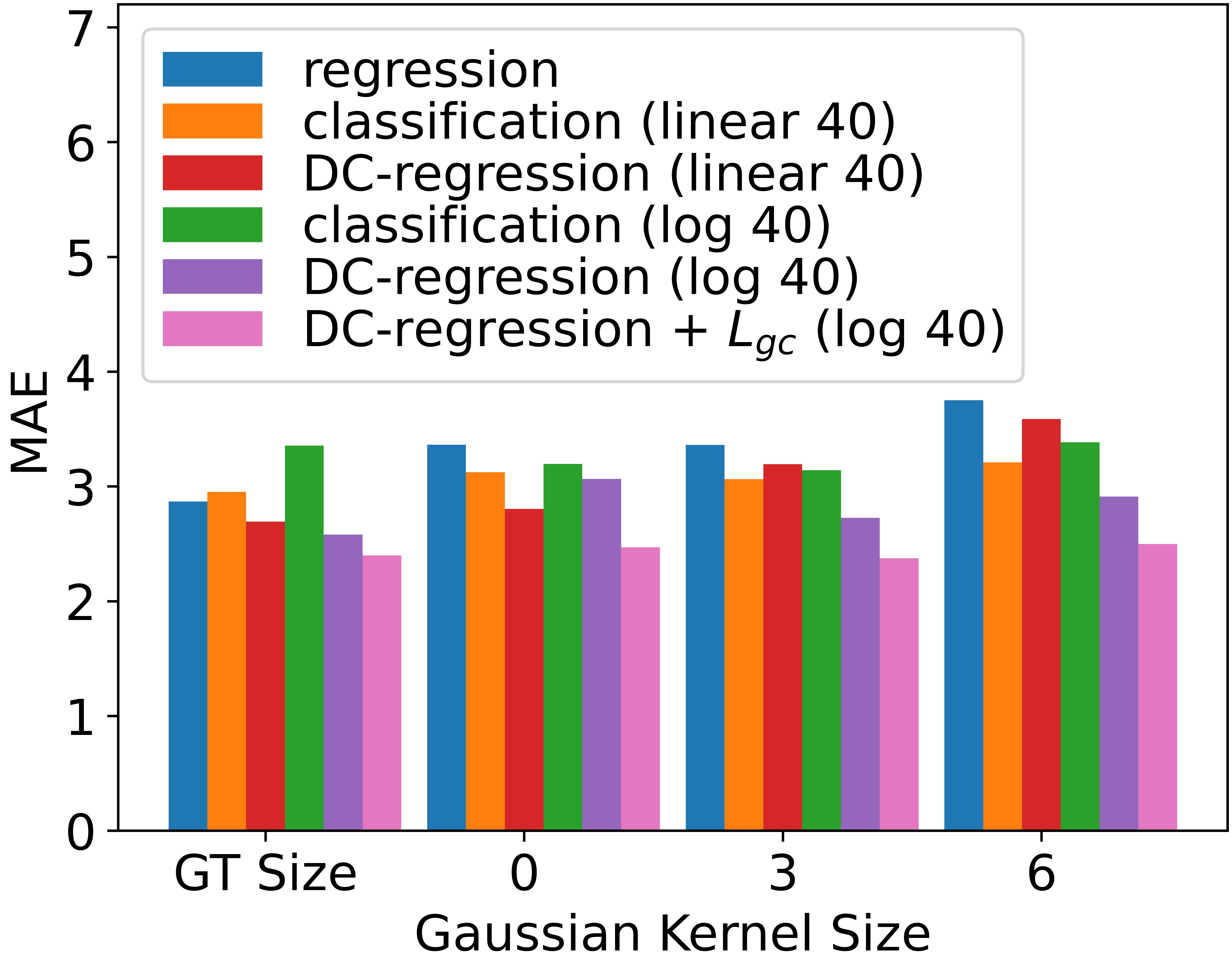}
\includegraphics[width=0.30\linewidth]{ 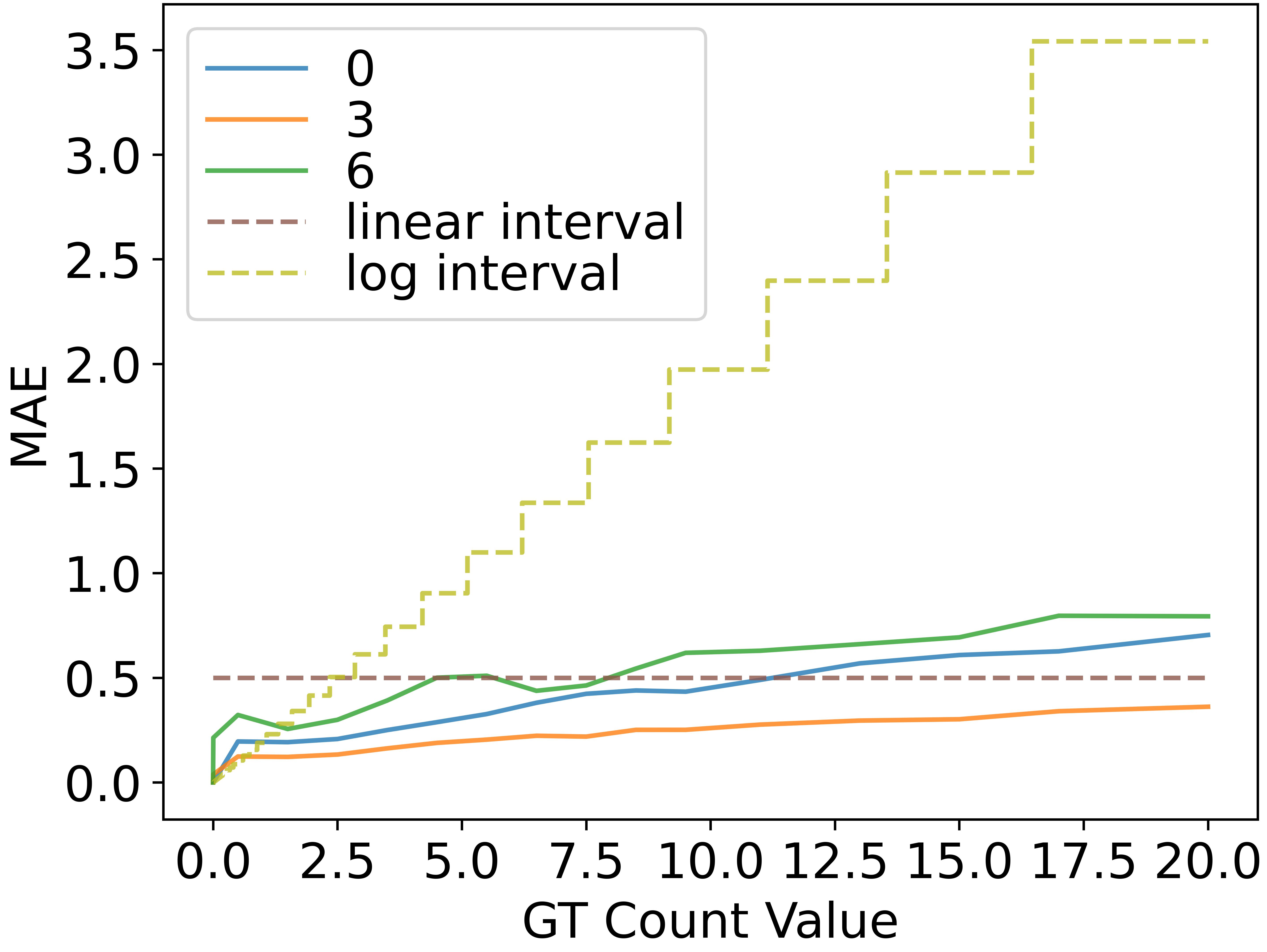}
\includegraphics[width=0.30\linewidth]{ 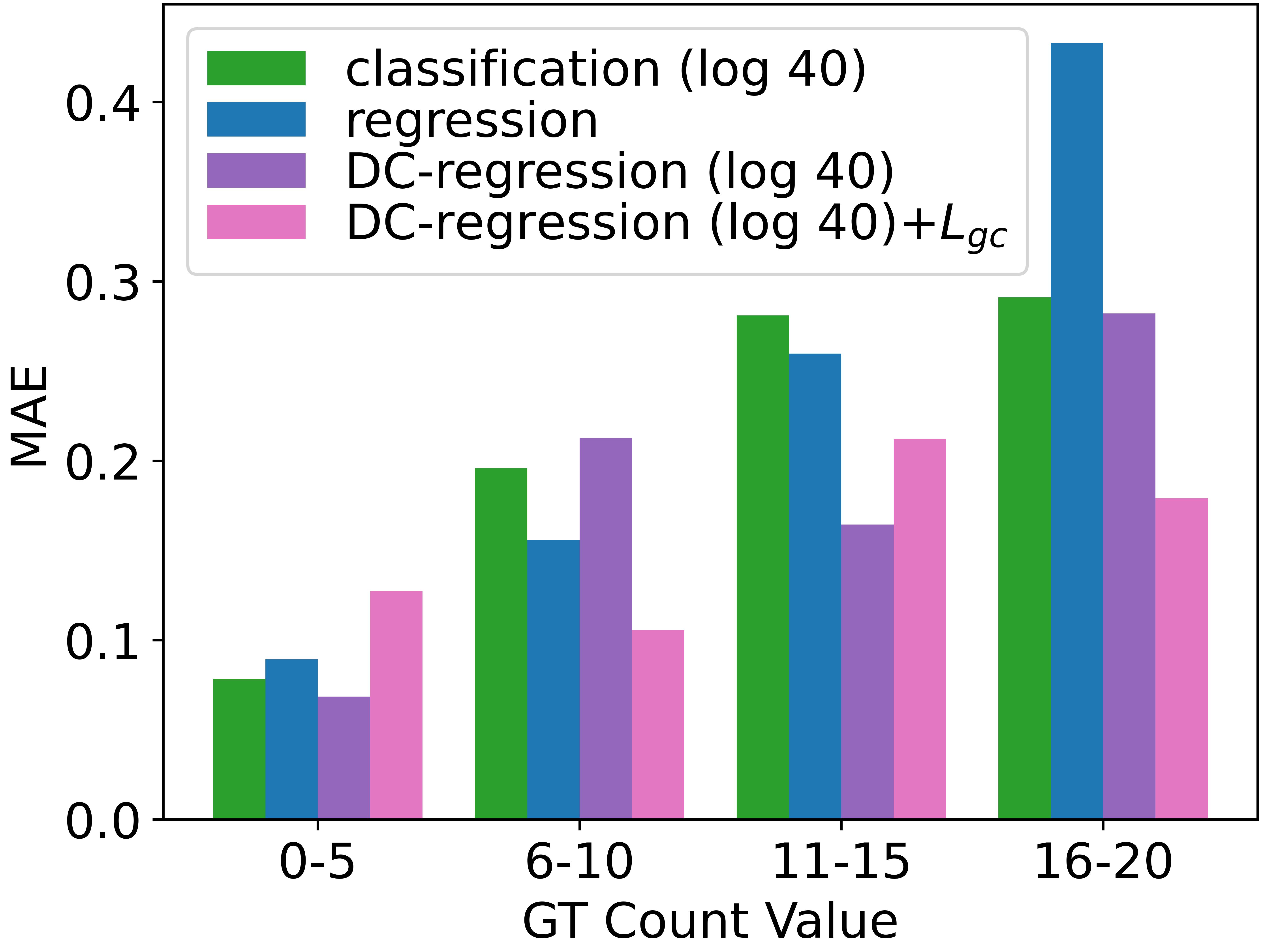}
\caption{Left: Counting results under various Gaussian kernel sizes. Here, $L_{\text{bias}}^{\lambda}$ is adopted as $L_{\text{gc}}$. Middle: The bias of local count value caused by various Gaussian kernel sizes.  Right: Counting error distribution w.r.t. local count value with $6$ as Gaussian kernel size. Linear/log denotes linear and log-spaced intervals respectively, while $20/40$ denotes the number of intervals.}
\label{fig:simu_gauss_size}
\end{figure}

add errors to the ground truth local counts. Fig.~\ref{fig:simu_gauss_size} shows that the error is monotonically  increasing with respect to $C$. Classification and DC-regression are more robust than regression under varied Gaussian kernel sizes.  Log intervals are better than linear intervals for classification when using a kernel of size 6, since the error of the ground truth is much larger than the linear interval length of $0.5$.  Finally, adding a global count loss $L_{bias}^{\lambda}$  into DC-regression decreases the count error in dense patches ($16\sim20$) where log intervals become too large.

\subsection{Study Findings}
We draw the following conclusions about local count models based on the observations from above:
\begin{enumerate}
    \item Partial objects make local counting harder.  All local count models (regression, classification and DC-regression) perform worse when partial objects are present in local image patches even if the ground truth local counts are perfectly generated according to object size.
    \item The ground truth local counts are not precise when partial objects are present and may contain error $\epsilon$. Both the bias of point annotation $(x_t,y_t)$ and Gaussian kernel $G_{\sigma}$ increase $\epsilon$ in local counts $C$. The error $\epsilon$ is monotonically increasing with respect to local count value $C$. As such, log-spaced intervals are more suitable than linear intervals, which use increasing interval length to handle increasing $\epsilon$.
    \item Classification and DC-regression perform much better than regression when $\epsilon$ is present, which suggests that it is beneficial to adopt discrete constraints in regression with imprecise ground truth. 
    \item Adding $L_{\text{gc}}$ into DC-regression effectively decreases discretization error in dense areas and improves the performance of DC-regression.
\end{enumerate}

\section{Experiment on Real-World Datasets}

\subsection{Datasets \& Implementation Details}

We verified the effectiveness of DC-regression on real-world counting and age estimation datasets. We evaluated DC-regression on three challenging crowd counting datasets (SHTech~\cite{MCNN_2016_CVPR}, JHU~\cite{sindagi2020jhu-crowd++} and QNRF~\cite{Compose_Loss_2018_ECCV}). Due to the varied size of images in counting datasets, we randomly cropped fixed-sized sub-images as training samples. The crop size for SHTech was $320\times320$, and $512\times 512$ for JHU and QNRF datasets.  For generating density maps, we followed the same settings as~\cite{MCNN_2016_CVPR,xiong2019open}.  To show the generality of DC-regression, we also evaluated it on two age estimation datasets, MegaAge (Mega) and MegaAsian (MegaA) datasets~\cite{zhang2017quantifying}. 

We followed the same architecture as specified in Sec.~\ref{sec:simulated}. We adopted the Adam optimizer, with a batch size of $8$ for crowd counting and $32$ for age estimation.  The initial learning rate of $10^{-4}$ was decreased by 0.1 whenever training error plateaued.  More implementation details are provided in the Supplementary.  For evaluation, we used $\text{MAE}$ to indicate counting accuracy and the Root Mean Squared Error ($\text{MSE}$) to reflect counting stability. Lower $\text{MAE}$ and $\text{MSE}$ indicate better counting performance.

\subsection{Ablation Study}

\subsubsection{Number of Intervals}
We adopted log-spaced intervals for both classification and DC-regression, varying the number of intervals $N$ from $2$ to $1000$. When $N=2$, there were only background $\{0\}$ and foreground classes $(0,V_{max}]$. For classification, we adopted the median value of the training samples within interval $(V_{i-1},V_{i}]$ to map the index back to a count. 

\begin{figure}[!t]
\centering
\includegraphics[width=0.40\linewidth]{ 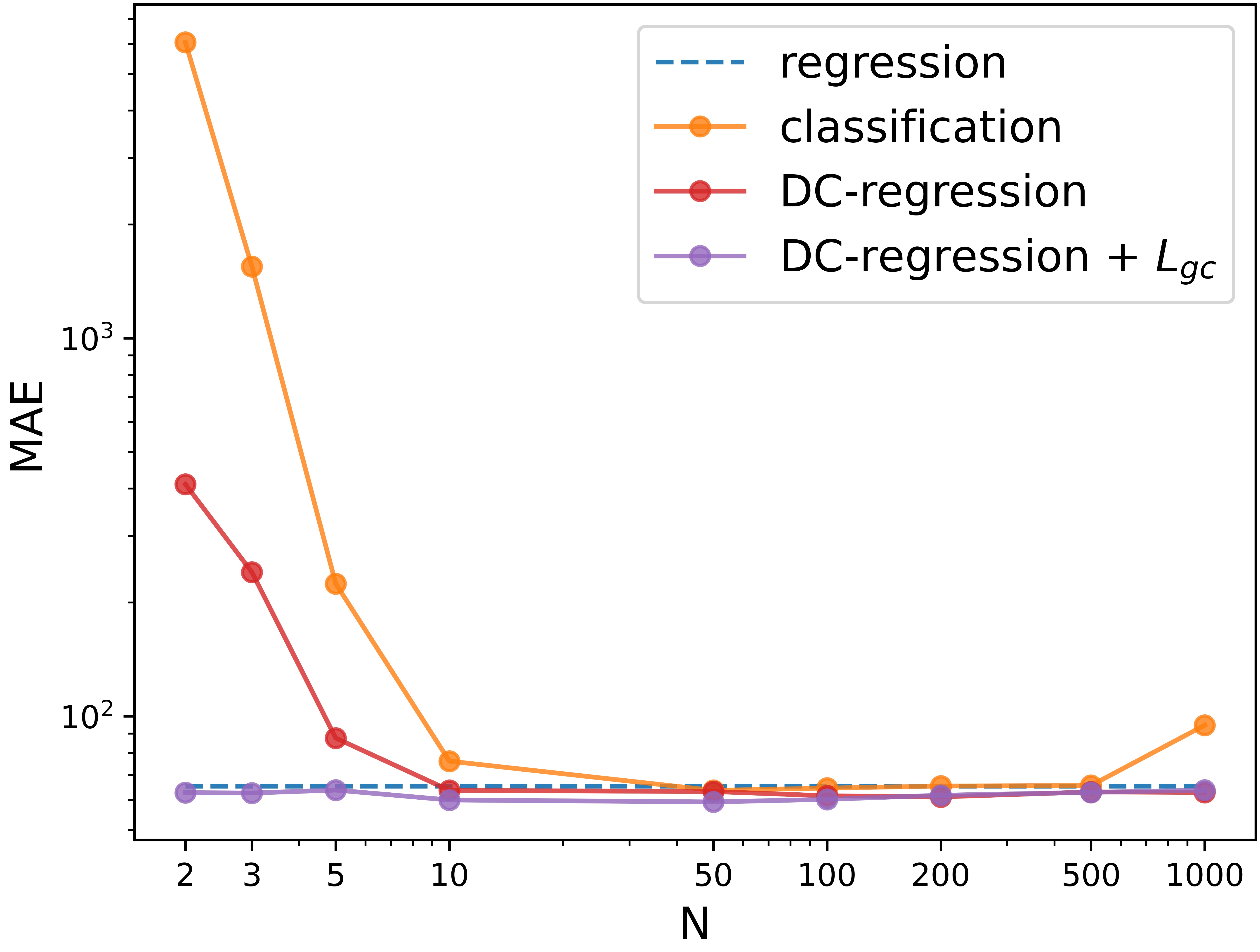}
\includegraphics[width=0.435\linewidth]{ 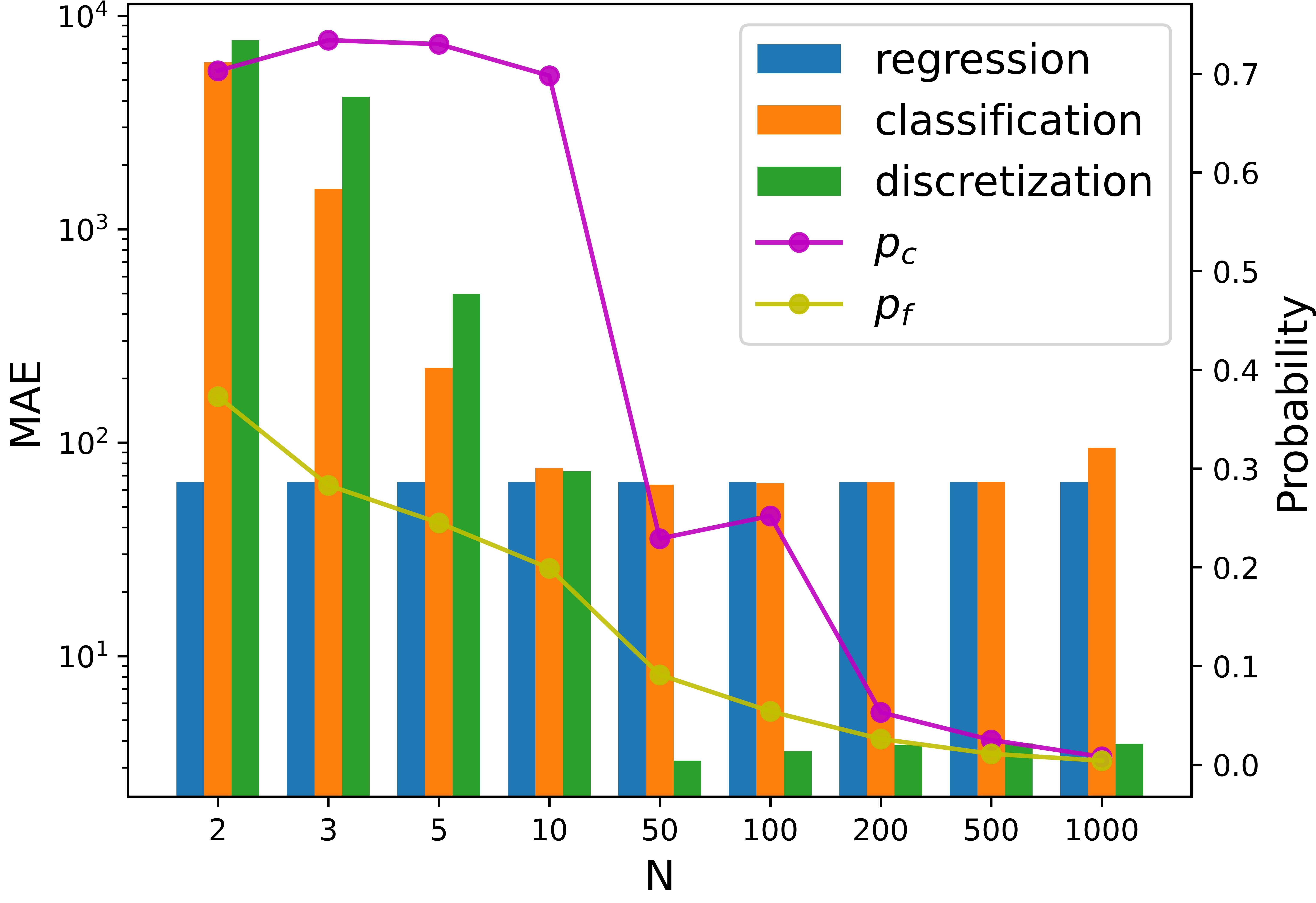}
\caption{The Effect of Number of Intervals ($N$) on on SHTech PartA dataset~\cite{MCNN_2016_CVPR}. Left: Counting error with respect to $N$. Right: MAE with classification probabilities ($p_c$ and $p_f$) with respect to $N$.  The discretization error (green bars) denotes the counting error of classification when all the class indexes are correctly predicted. $p_c$ and $p_f$ denotes the average probability of correctly or falsely predicted samples, respectively. $L_{\text{bias}}^{\lambda}$ is adopted as $L_{\text{gc}}$.}
\label{fig:class_number}
\end{figure}

From Fig.~\ref{fig:class_number}, we observe that regression sets a baseline performance indicated by the dashed line.  With the right selection of $N$, all the discrete models surpass the regression baseline, but a poor selection of $N$ hurts the performance.  For classification, when $N$ is small, discretization error of the intervals increases the MAE.  When $N$ is large, MAE increases again as the classes are no longer distinguishable. 
DC-regression also shows poor performance when $N<=5$ as it is similarly affected by discretization errors like classification.  However, it has stable performance when $N>5$ as it converges to standard regression when $N\rightarrow \infty$. The global loss $L_{\text{gc}}$ mitigates the discretization errors of DC-regression when $N<=5$, making this the optimal combination.  As classification showed the best performance when $N\!=\!100$, we keep $N\!=\!100$ for the remaining experiments.

\subsubsection{Interval Spacing}

We compared linear, log-spaced and uep~\cite{wang2021uniformity} intervals for classification and DC-regression. $L_{\text{bias}}^{\lambda}$ from Eq.~\eqref{eq:L_bias_lambda} is chosen as $L_{\text{gc}}$. Table~\ref{tab:interval_types} shows that log-spaced intervals, in line with the synthetic experiments, are better than linear intervals for both classification  and DC-regression.  After adding the global count loss $L_{\text{gc}}$ to DC-regression, counting performance improves and all three types of intervals perform similarly. This confirms that $L_{\text{gc}}$ mitigates discretization errors and makes DC-regression less sensitive to interval spacing.

\begin{table}[!t] 
\caption{Comparison different kind of interval partitions in discrete counting models}
\label{tab:interval_types}
\centering
\begin{tabular}{l|l|ll|ll}
\hline
       \multirow{2}{*}{Method}             &    \multirow{2}{*}{Interval Partition }          & \multicolumn{2}{c|}{SHA} & \multicolumn{2}{c}{SHB}  \\ \cline{3-6}
            &  & MAE & MSE & MAE & MSE\\ \hline

 \multirow{2}{*}{classification}                   & linear         & 65.6   & 115.4 & 8.6   & 14.6                                         \\
     & log            & 64.3  & 112.3 & 8.7   & 17.2                                         \\ 
                   & uep~\cite{wang2021uniformity}            & 63.9  & 112.8 & 7.9   & 15.1  \\
\hline

\multirow{2}{*}{DC-regression} & linear & 62.5  & 106.0  & 7.5   & 12.2                                          \\
 & log            & 61.6  & 96.7  & 7.1   & 11.1 \\ 
                     & uep~\cite{wang2021uniformity}           & 61.9  & 104.2 & 7.4   & 12.2                                          \\                 \hline                 
    
\multirow{2}{*}{DC-regression+$L_{\text{gc}}$} & linear & \textbf{60.3}  &  \textbf{95.5} & \textbf{6.6}   &  11.0                                         \\
 & log            & \textbf{60.3}  & 103.7  &  6.7  & 10.6 \\
 & uep~\cite{wang2021uniformity}           & 61.3  & 97.5 & 6.7   & \textbf{10.2}                                          \\

 \hline
\end{tabular}
\end{table}

\subsubsection{Choice of Global Count Loss $L_{\text{gc}}$}
Table~\ref{tab:Lgc} compares the different global count loss terms. When adding global count losses $L_{BL}$ from~\cite{ma2019bayesian} and $L_{\text{bias}}^{\lambda}$ to regression, the performance surpasses standard regression and classification with $L_1$ or cross entropy loss $L_{cls}$.  However, it does not surpass standard DC-regression, suggesting that local count supervision is still more effective than global count supervision under discrete constraints. When looking at DC-regression specifically, the naive global count $L_{\text{c}}$ from Eq.~\eqref{eq:L_count} harms DC-regression, as explained in Sec.~\ref{subsec:L_gc_define}.  However, being selective on the local patch for the global count rectifies this error, with $L_{\text{bias}}^{\lambda}$ being more effective than $L_{\text{bias}}^{0}$, which shows no improvement. We refer the reader to the Supplementary for a detailed comparison and discussion between $L_{\text{bias}}^{0}$ and $L_{\text{bias}}^{\lambda}$.  We further observe that $L_{BL}$ is also helpful for discrete regression and shows comparable results as $L_{\text{bias}}^{\lambda}$.

\begin{table}[!t]
\caption{Comparison of different $L_{\text{gc}}$. $L_{cls}$ denotes the standard cross-entropy loss used in classification.}
\label{tab:Lgc}
\centering
\begin{tabular}{l|c|c|ll|ll}
\hline
&\multirow{2}{*}{main loss}&\multirow{2}{*}{$L_{\text{gc}}$} & \multicolumn{2}{c|}{SHA} & \multicolumn{2}{c}{SHB}\\
\cline{4-7}
 && &MAE        & MSE        & MAE        & MSE \\
\hline
regression&$L_{1}$&---& 65.4      & 103.3    & 10.7      & 19.5     \\
\hline
classification&$L_{cls}$&---& 64.6     & 106.7    & 8.7      & 17.2     \\
\hline
\multirow{2}{*}{regression + $L_{\text{gc}}$}&---&$L_{BL}$&62.1&103.4&7.4&10.8\\
&---&$L_{\text{bias}}^{\lambda}$&62.9&108.5&7.8&12.0\\

\hline
&$L_{dc}$&---&61.6 & \textbf{96.7}  & 7.1 & 11.1   \\
&$L_{dc}$&$L_{\text{c}}$&63.5 & 104.4 & 7.5 & 13.2   \\
DC-regression&$L_{dc}$&$L_{\text{bias}}^{0}$&61.6 & 105.3 & 7.1 & 11.9  \\
&$L_{dc}$&$L_{\text{bias}}^{\lambda}$&\textbf{60.3} & 103.7 & \textbf{6.7} & \textbf{10.6} \\
&$L_{dc}$&$L_{BL}$&60.7&101.0&7.1&11.0\\
\hline
\end{tabular}
\end{table}

\subsection{Adding DC-Regression to State-of-the-Art}
S-DCNet, proposed by Xiong~\textit{et al.}~\cite{xiong2019open}, is a classification-based state-of-the-art local counting model.  As it was proposed to tackle open-set counting, 
it features two loss functions: $L_c$ for supervising closed-set counters and $L_m$ for supervising local counts outside the closed set.  We replaced these losses by adopting DC-regression's $L_{dc}$ from Eq.~\eqref{eq:ldc} for $L_c$ and $L_{dc}+L_{\text{gc}}$ for $L_m$.  Note that we do not consider the global count loss for closed set counting as the closed counters only predict truncated counts outside the closed set range. 

We also added a standard S-DCNet regression baseline (`reg') for comparison. Table~\ref{tab:sdcnet} shows that S-DCNet (reg) performs worse than the classification variant of S-DCNet (`cls') and is consistent with the conclusion in~\cite{xiong2019open}. The DC-Regression variant of S-DCNet (`dcreg'), however, is comparable or better than S-DCNet (cls), verifying the effectiveness of DC-regression.  

\begin{table}[!t]
\caption{Applying DC-regression in S-DCNet~\cite{xiong2019open} }
\label{tab:sdcnet}
\centering
\begin{tabular}{l|l|l|ll|ll|ll|ll}

\hline
           &\multirow{2}{*}{$L_c$} &\multirow{2}{*}{$L_m$}    & \multicolumn{2}{c|}{SHA} & \multicolumn{2}{c|}{SHB} &  \multicolumn{2}{c|}{JHU}&\multicolumn{2}{c}{QNRF}  \\
\cline{4-11}
              & && MAE        & MSE        & MAE        & MSE        & MAE         & MSE        & MAE        & MSE        \\
\hline

S-DCNet (cls)~\cite{xiong2019open}&$L_{cls}$&$L_1$    & \textbf{58.3}       & 95.0         & \textbf{6.7}        & \textbf{10.7}             & 65.2       & 272.8  & 104.4       & 176.1 \\
S-DCNet (reg)&$L_1$&$L_1$+$L_{BL}$ & 61.1       & 94.2       & 7.4        & 12.5             & 66.1       & 272.1 & 92.3        & 158.8      \\
S-DCNet (dcreg)$^{*}$&$L_{dc}$&$L_1$+$L_{BL}$  & 59.7       & \textbf{91.4}       & 7.0        & 11.6            & \textbf{60.0}       & 269.9    & 86.9        & 159.3   \\
S-DCNet (dcreg)$^{\dag}$&$L_{dc}$&$L_1$+$L_{\text{bias}}^{\lambda}$ &59.8&100.0&6.8&11.5&62.1&\textbf{268.9}&\textbf{84.8}&\textbf{142.3}\\
\hline
\end{tabular}
\end{table}

\subsection{Comparison on Crowd Counting Datasets}

We compare DC-regression with other state-of-the-art counting methods on three crowd counting datasets in Table~\ref{tab:comp}.  DC-regression outperforms local count regression and classification on all the datasets.  Adding global count loss such as $L_{BL}$ or  $L_{\text{bias}}^{\lambda}$ further improves the results, suggesting that a global constraint is helpful for local count models.  In particular, DC-regression and S-DCNet (dcreg) show comparable performance with state-of-the-art approaches. Specifically, our methods are better than or comparable with density regression methods~\cite{wang2020DMCount,ma2021learning,wan2021generalized}, which adopt optimum transportation~\cite{peyre2019computational} to model the imperfect ground truth of density maps.

\begin{table}[!t]
\caption{Comparison with State-of-the-art method on Crowd Counting Datasets. Methods are grouped as density map regression, local count regression, classification and DC-Regression approaches}
\label{tab:comp}
\centering
\begin{tabular}{l|l|ll|ll|ll|ll}
\hline
 & \multirow{2}{*}{Backbone}  & \multicolumn{2}{c|}{SHA}   & \multicolumn{2}{c|}{SHB} & \multicolumn{2}{c|}{JHU}  & \multicolumn{2}{c}{QNRF}                             \\
  \cline{3-10}
 &   & MAE & MSE & MAE & MSE & MAE & MSE & MAE & MSE  \\
  \hline
CSRNet~\cite{CSRNet_2018_CVPR}  & VGG16   &68.2&115.0  &10.6&16.0    &85.9&309.2 &108.2&	 181.3   \\
DRCN~\cite{sindagi2020jhu-crowd++} &VGG16 &64.0&98.4 &8.5&14.4  &82.3&328.0&112.2&176.3\\
BL~\cite{ma2019bayesian} &VGG19 &62.8&101.8 &7.7&12.7 &75.0&299.9 &88.7&154.8 \\
PaDNet~\cite{tian2019padnet}&VGG16& 59.2& 98.1& 8.1& 12.2&--- &---&96.5 &170.2\\
MNA~\cite{wan2020modeling}&VGG19&61.9 &99.6& 7.4& 11.3&67.7&258.5 &85.8& 150.6\\
OT~\cite{wang2020DMCount}&VGG19&59.7&95.7&7.4&11.8&68.4 &283.3&85.6&148.3\\
UOT~\cite{ma2021learning}&VGG19&\textbf{58.1}&95.9&\textbf{6.5}&\textbf{10.2}&60.5&\textbf{252.7}&\textbf{83.3}&\textbf{142.3}\\
Generalized Loss~\cite{wan2021generalized}&VGG19&61.3&95.4&7.3&11.7&\textbf{59.9}&259.5&84.3&147.5\\

\hline
regression ($L_{reg}$)              & VGG16  & 65.4                   & 103.3                & 10.7                   & 19.5             & 71.2                  & 296.0      & 98.6                  & 166.6                                 \\
$L_{BL}$& VGG16 & 62.2                  & 103.4                  & 7.4                    & 10.7                & 64.2                  & 275.7  & 90.1                  & 162.5                                 \\
$L_{\text{bias}}^{\lambda}$ & VGG16                     & 62.9                    & 108.5                 & 7.8                  & 12.0               & 68.6                  & 289.4   & 93.3                  & 160.8                                  \\
\hline
classification    & VGG16        & 64.6                  & 106.7                 & 8.7                   & 17.2            & 67.8                  & 261.6       & 97.6                  & 163.2                                \\

S-DCNet (cls)~\cite{xiong2019open} & VGG16  & 58.3       & 95.0         & 6.7        & 10.7             & 65.2       & 272.8  & 104.4       & 176.1 \\
\hline
DC-regression  & VGG16     & 61.6                 & 96.7                 & 7.1                   & 11.1           & 67.2                 & 288.2       & 91.4                  & 157.5                                  \\
DC-regression+$L_{\text{bias}}^{\lambda}$& VGG16 & 60.3                 & 103.7                 & 6.7                   & 10.6              & 64.8                  & 282.6     & 86.0                  & 148.2                                \\
DC-regression+$L_{BL}$  & VGG16 & 60.7                  & 101.0                & 7.1                   & 11.0          & 61.6                  & 263.2        & 87.1                  & 152.1                             \\
S-DCNet (dcreg)$^{*}$ & VGG16 & 59.7       & \textbf{91.4}       & 7.0        & 11.6            & 60.0       & 269.9    & 86.9        & 159.3   \\
S-DCNet (dcreg)$^{\dag}$ & VGG16 &59.8&100.0&6.8&11.5&62.1&268.9&84.8&\textbf{142.3}\\
\hline
\end{tabular}
\end{table}

\subsection{Comparison on Age Estimation Datasets}

Unlike local counts, human age is modelled as integer counts in standard age-estimation datasets~\cite{zhang2017quantifying}.  As such, we postulate that applying our discrete constraint to regress age may also be suitable.  We verify the effectiveness of DC-regression on age estimation datasets Mega and MegaA~\cite{zhang2017quantifying}, adopting a linearly spaced interval of length $1$ as age increases with step $1$. To evaluate, we use MAE, MSE and CAi (i=3,5,7), where CAi denotes the proportion of samples with MAE less than $i$. Table~\ref{tab:intra_age} shows that DC-regression is better than regression, which suggests that it is better to use discrete constraints for age prediction. Similar to the analysis of counting, we should ignore the loss of age when the prediction is within the class intervals, in order to provide the correct gradient to benefit the training process.

\begin{table*}[!t] 
	\caption{Comparison on age prediction datasets}
	\label{tab:intra_age}
	\centering
	\begin{tabular}{l|lllll|lllll}
		\hline
		\multirow{2}{*}{Method}     &  \multicolumn{5}{c|}{Mega} & \multicolumn{5}{c}{MegaA} \\
		\cline{2-11}
		&MAE&RMSE&CA3&CA5&CA7 &MAE&RMSE&CA3&CA5&CA7\\
		\hline
		Posterior~\cite{zhang2017quantifying} 
		& \ --- \ & \ --- \ & 38.69 & 57.90 & 73.15 &\ --- \ & \ --- \ & 62.08 & 80.43 & 90.42\\ 

		Xia \textit{et~al.}~\cite{xia2020multi}  
		&  \ --- \ & \ --- \ & \ --- \ & \ --- \ & \ --- \ & 2.80 & \ --- \ & 62.50 & 82.37 & \ --- \ \\ \
		Yu \textit{et~al.}~\cite{tingting2019three} 
		&  \ --- \ & \ --- \ & 42.19 & 60.0 & 72.70 & \ --- \ & \ --- \ & 64.80 & 83.20 & 91.40 \\

\hline
classification & 5.57 & 7.15 & 39.72 & 57.10 & 71.45 & 2.91  & 4.14 & 68.19 & 84.82 & 93.03 \\
regression     & 5.26 & 6.72 & 41.89 & 59.84 & 74.73 & 2.87 & 4.00 & 68.57 & 85.10 & 93.51 \\
DC-regression           & \textbf{5.15} & \textbf{6.58}  & \textbf{42.36} & \textbf{61.31} & \textbf{75.26} & \textbf{2.80} & \textbf{3.97}  & \textbf{69.40} & \textbf{85.98} & \textbf{93.79} \\
		\hline
	\end{tabular}
\end{table*}

\section{Conclusion}
In this paper, we experimentally showed that ground truth local counts are error-prone, and classification outperforms regression when local counts are imprecise. The disadvantage of regression could be mitigated by adopting discrete constraints. We proposed DC-regression to handle the ground truth error in local count models. DC-regression showed superior results in counting tasks compared to classification and regression, and it is also suitable for age estimation tasks. 

\paragraph{Acknowledgments} This research is supported by the Ministry of Education, Singapore, under its MOE Academic Research Fund Tier 2 (STEM RIE2025 MOE-T2EP20220-0015).

\bibliographystyle{splncs04}


\begin{thebibliography}{10}\itemsep=-1pt

\bibitem{gu2022dive}
Kerui Gu, Linlin Yang, and Angela Yao.
\newblock Dive deeper into integral pose regression.
\newblock In {\em International Conference on Learning Representations (ICLR)},
  2022.

\bibitem{Compose_Loss_2018_ECCV}
Haroon Idrees, Muhmmad Tayyab, Kishan Athrey, Dong Zhang, Somaya Al-Maadeed,
  Nasir Rajpoot, and Mubarak Shah.
\newblock Composition loss for counting, density map estimation and
  localization in dense crowds.
\newblock In {\em The European Conference on Computer Vision (ECCV)}, pages
  532--546, 2018.

\bibitem{lempitsky2010learning}
Victor Lempitsky and Andrew Zisserman.
\newblock Learning to count objects in images.
\newblock {\em Advances in neural information processing systems}, 23, 2010.

\bibitem{li2018deep}
Ruibo Li, Ke Xian, Chunhua Shen, Zhiguo Cao, Hao Lu, and Lingxiao Hang.
\newblock Deep attention-based classification network for robust depth
  prediction.
\newblock In {\em Proc. Asian Conference on Computer Vision (ACCV)}, 2018.

\bibitem{CSRNet_2018_CVPR}
Yuhong Li, Xiaofan Zhang, and Deming Chen.
\newblock Csrnet: Dilated convolutional neural networks for understanding the
  highly congested scenes.
\newblock In {\em The IEEE Conference on Computer Vision and Pattern
  Recognition (CVPR)}, pages 1091--1100, 2018.

\bibitem{liu2019counting}
Liang Liu, Hao Lu, Haipeng Xiong, Ke Xian, Zhiguo Cao, and Chunhua. Shen.
\newblock Counting objects by blockwise classification.
\newblock {\em IEEE Trans. on Circuits and Systems for Video Technology},
  30(10):3513--3527, 2019.

\bibitem{liu2020adaptive}
Xiyang Liu, Jie Yang, Wenrui Ding, Tieqiang Wang, Zhijin Wang, and Junjun
  Xiong.
\newblock Adaptive mixture regression network with local counting map for crowd
  counting.
\newblock In {\em European Conference on Computer Vision}, pages 241--257.
  Springer, 2020.

\bibitem{ma2019bayesian}
Zhiheng Ma, Xing Wei, Xiaopeng Hong, and Yihong Gong.
\newblock Bayesian loss for crowd count estimation with point supervision.
\newblock In {\em Proceedings of the IEEE/CVF International Conference on
  Computer Vision}, pages 6142--6151, 2019.

\bibitem{ma2021learning}
Zhiheng Ma, Xing Wei, Xiaopeng Hong, Hui Lin, Yunfeng Qiu, and Yihong Gong.
\newblock Learning to count via unbalanced optimal transport.
\newblock In {\em Proceedings of the AAAI Conference on Artificial
  Intelligence}, pages 2319--2327, 2021.

\bibitem{newell2016stacked}
Alejandro Newell, Kaiyu Yang, and Jia Deng.
\newblock Stacked hourglass networks for human pose estimation.
\newblock In {\em European Conference on Computer Vision (ECCV)}, 2016.

\bibitem{onoro2016towards}
Daniel Onoro-Rubio and Roberto~J L{\'o}pez-Sastre.
\newblock Towards perspective-free object counting with deep learning.
\newblock In {\em European conference on computer vision}, pages 615--629.
  Springer, 2016.

\bibitem{paul2017count}
Joseph Paul~Cohen, Genevieve Boucher, Craig~A Glastonbury, Henry~Z Lo, and
  Yoshua Bengio.
\newblock Count-ception: Counting by fully convolutional redundant counting.
\newblock In {\em Proceedings of the IEEE International conference on computer
  vision workshops}, pages 18--26, 2017.

\bibitem{peyre2019computational}
Gabriel Peyr{\'e}, Marco Cuturi, et~al.
\newblock Computational optimal transport: With applications to data science.
\newblock {\em Foundations and Trends{\textregistered} in Machine Learning},
  11(5-6):355--607, 2019.

\bibitem{shi2019counting}
Zenglin Shi, Pascal Mettes, and Cees~GM Snoek.
\newblock Counting with focus for free.
\newblock In {\em Proceedings of the IEEE/CVF International Conference on
  Computer Vision}, pages 4200--4209, 2019.

\bibitem{Simonyan2014Very_VGG16}
Karen Simonyan and Andrew Zisserman.
\newblock Very deep convolutional networks for large-scale image recognition.
\newblock {\em Computer Science}, 2014.

\bibitem{sindagi2019multi}
Vishwanath~A Sindagi and Vishal~M Patel.
\newblock Multi-level bottom-top and top-bottom feature fusion for crowd
  counting.
\newblock In {\em Proceedings of the IEEE/CVF international conference on
  computer vision}, pages 1002--1012, 2019.

\bibitem{sindagi2020jhu-crowd++}
Vishwanath~A Sindagi, Rajeev Yasarla, and Vishal~M Patel.
\newblock Jhu-crowd++: Large-scale crowd counting dataset and a benchmark
  method.
\newblock {\em Technical Report}, 2020.

\bibitem{tian2019padnet}
Yukun Tian, Yiming Lei, Junping Zhang, and James~Z Wang.
\newblock Padnet: Pan-density crowd counting.
\newblock {\em IEEE Transactions on Image Processing}, 29:2714--2727, 2019.

\bibitem{tingting2019three}
Yu Tingting, Wang Junqian, Wu Lintai, and Xu Yong.
\newblock Three-stage network for age estimation.
\newblock {\em CAAI Transactions on Intelligence Technology}, 4(2):122--126,
  2019.

\bibitem{wan2020modeling}
Jia Wan and Antoni Chan.
\newblock Modeling noisy annotations for crowd counting.
\newblock {\em Advances in Neural Information Processing Systems}, 33, 2020.

\bibitem{wan2021generalized}
Jia Wan, Ziquan Liu, and Antoni~B Chan.
\newblock A generalized loss function for crowd counting and localization.
\newblock In {\em Proceedings of the IEEE/CVF Conference on Computer Vision and
  Pattern Recognition}, pages 1974--1983, 2021.

\bibitem{wang2020DMCount}
Boyu Wang, Huidong Liu, Dimitris Samara, and Minh Hoai.
\newblock Distribution matching for crowd counting.
\newblock In {\em Conference on Neural Information Processing Systems
  (NeurIPS)}, 2020.

\bibitem{wang2021uniformity}
Changan Wang, Qingyu Song, Boshen Zhang, Yabiao Wang, Ying Tai, Xuyi Hu,
  Chengjie Wang, Jilin Li, Jiayi Ma, and Yang Wu.
\newblock Uniformity in heterogeneity: Diving deep into count interval
  partition for crowd counting.
\newblock In {\em Proceedings of the IEEE/CVF International Conference on
  Computer Vision}, pages 3234--3242, 2021.

\bibitem{xia2020multi}
Min Xia, Xu Zhang, Liguo Weng, Yiqing Xu, et~al.
\newblock Multi-stage feature constraints learning for age estimation.
\newblock {\em IEEE Transactions on Information Forensics and Security},
  15:2417--2428, 2020.

\bibitem{xiao2018simple}
Bin Xiao, Haiping Wu, and Yichen Wei.
\newblock Simple baselines for human pose estimation and tracking.
\newblock In {\em European Conference on Computer Vision}, pages 466--481,
  2018.

\bibitem{xiong2019tasselnetv2}
Haipeng Xiong, Zhiguo Cao, Hao Lu, Simon Madec, Liang Liu, and Chunhua Shen.
\newblock Tasselnetv2: in-field counting of wheat spikes with context-augmented
  local regression networks.
\newblock {\em Plant Methods}, 15(1):1--14, 2019.

\bibitem{xiong2019open}
Haipeng Xiong, Hao Lu, Chengxin Liu, Liang Liu, Zhiguo Cao, and Chunhua Shen.
\newblock From open set to closed set: Counting objects by spatial
  divide-and-conquer.
\newblock In {\em Proceedings of the IEEE/CVF International Conference on
  Computer Vision}, pages 8362--8371, 2019.

\bibitem{zhang2015cross}
C. Zhang, H. Li, X. Wang, and X. Yang.
\newblock Cross-scene crowd counting via deep convolutional neural networks.
\newblock In {\em CVPR}, 2015.

\bibitem{zhang2017quantifying}
Yunxuan Zhang, Li Liu, Cheng Li, et~al.
\newblock Quantifying facial age by posterior of age comparisons.
\newblock {\em arXiv preprint arXiv:1708.09687}, 2017.

\bibitem{MCNN_2016_CVPR}
Yingying Zhang, Desen Zhou, Siqin Chen, Shenghua Gao, and Yi Ma.
\newblock Single-image crowd counting via multi-column convolutional neural
  network.
\newblock In {\em The IEEE Conference on Computer Vision and Pattern
  Recognition (CVPR)}, pages 589--597, 2016.

\end{thebibliography}

\end{document}